\documentclass[journal,twoside,web]{ieeecolor}
\usepackage{generic}
\usepackage{cite}
\usepackage{amsmath,amssymb,amsfonts}
\usepackage{algorithmic}
\usepackage{graphicx}
\usepackage{algorithm,algorithmic}
\usepackage{hyperref}
\hypersetup{hidelinks=true}
\usepackage{textcomp}
\usepackage{multirow}
\usepackage{colortbl}
\providecommand{\refname}{REFERENCES}
\def\BibTeX{{\rm B\kern-.05em{\sc i\kern-.025em b}\kern-.08em
    T\kern-.1667em\lower.7ex\hbox{E}\kern-.125emX}}
\begin{document}
\bstctlcite{IEEEexample:BSTcontrol}
\title{SurgNarrator: A Generative Retrieval Framework for Surgical Video Understanding}
\author{Yuqing Feng, Jiawei Ma, Kevin Qinghong Lin, Kun Yuan, Nicolas Padoy, Daniel S. Elson, Anh Nguyen, \IEEEmembership{Member, IEEE}, Stamatia Giannarou, Baoru Huang, \IEEEmembership{Member, IEEE}
\thanks{Y. Feng, A. Nguyen, and B. Huang are with the Department of
Computer Science, University of Liverpool, Liverpool L69 3BX, U.K.
(e-mail: Yuqing.Feng@liverpool.ac.uk; Anh.Nguyen@liverpool.ac.uk; Baoru.Huang@liverpool.ac.uk).}
\thanks{J. Ma is with the Department of Computer Science, City University
of Hong Kong, Hong Kong SAR, China (e-mail: jiaweima@cityu.edu.hk).}
\thanks{K. Q. Lin and P. Torr are with the Department of Engineering,
University of Oxford, Oxford OX1 3PJ, U.K.
(e-mail: kevin.qh.lin@gmail.com; philip.torr@eng.ox.ac.uk).}
\thanks{K. Yuan and N. Padoy are with the University of Strasbourg, CNRS, INSERM, ICube, UMR7357, and with IHU Strasbourg, Strasbourg, France (e-mail: kun.yuan@ext.ihu-strasbourg.eu; npadoy@unistra.fr).}
\thanks{D. S. Elson and S. Giannarou are with the Hamlyn Centre for Robotic Surgery,
Department of Surgery and Cancer, Imperial College London, London SW7 2AZ,
U.K. (e-mail: daniel.elson@imperial.ac.uk; stamatia.giannarou@imperial.ac.uk).}
}

\maketitle

\begin{abstract}
Surgical procedures unfold as structured and recurring clinical events, whose real-time understanding via intraoperative surgical videos is critical for intraoperative decision-making and support. However, existing video understanding methods force a trade-off: autoregressive video-language models support comprehensive reasoning but are not practical for time-sensitive clinical applications, whereas contrastive models offer low latency but struggle with complex scene understanding. Recently, generative retrieval has been explored for general-domain video understanding, but transferring it to surgery is not trivial because near-identical visual appearances may indicate semantically distinct events, and the terminology involved is highly surgery-specific. To this end, we propose SurgNarrator, a new generative retrieval framework tailored for surgical video understanding. We construct a well-curated surgery-centric vocabulary from surgical captions to define a clinically meaningful retrieval space. We then adapt the pre-trained Qwen3-VL-Embedding-8B to learn discriminative clinical representations with a temporally-aware contrastive objective. During inference, a hierarchical, procedure-aware retrieval strategy narrows the search space to the relevant procedure type, delivering fast and effective responses. Our method is comprehensively evaluated on twelve benchmarks in a zero-shot setting and achieves consistent performance gains over state-of-the-art baselines, while reducing output-stage latency by more than two orders of magnitude compared with the generative baseline.
\end{abstract}

\begin{IEEEkeywords}
Contrastive learning, surgical video understanding, video-language model, video-text retrieval.
\end{IEEEkeywords}

\section{Introduction}
\label{sec:introduction}

\begin{figure*}[!t]
\centering
\includegraphics[width=1\textwidth]{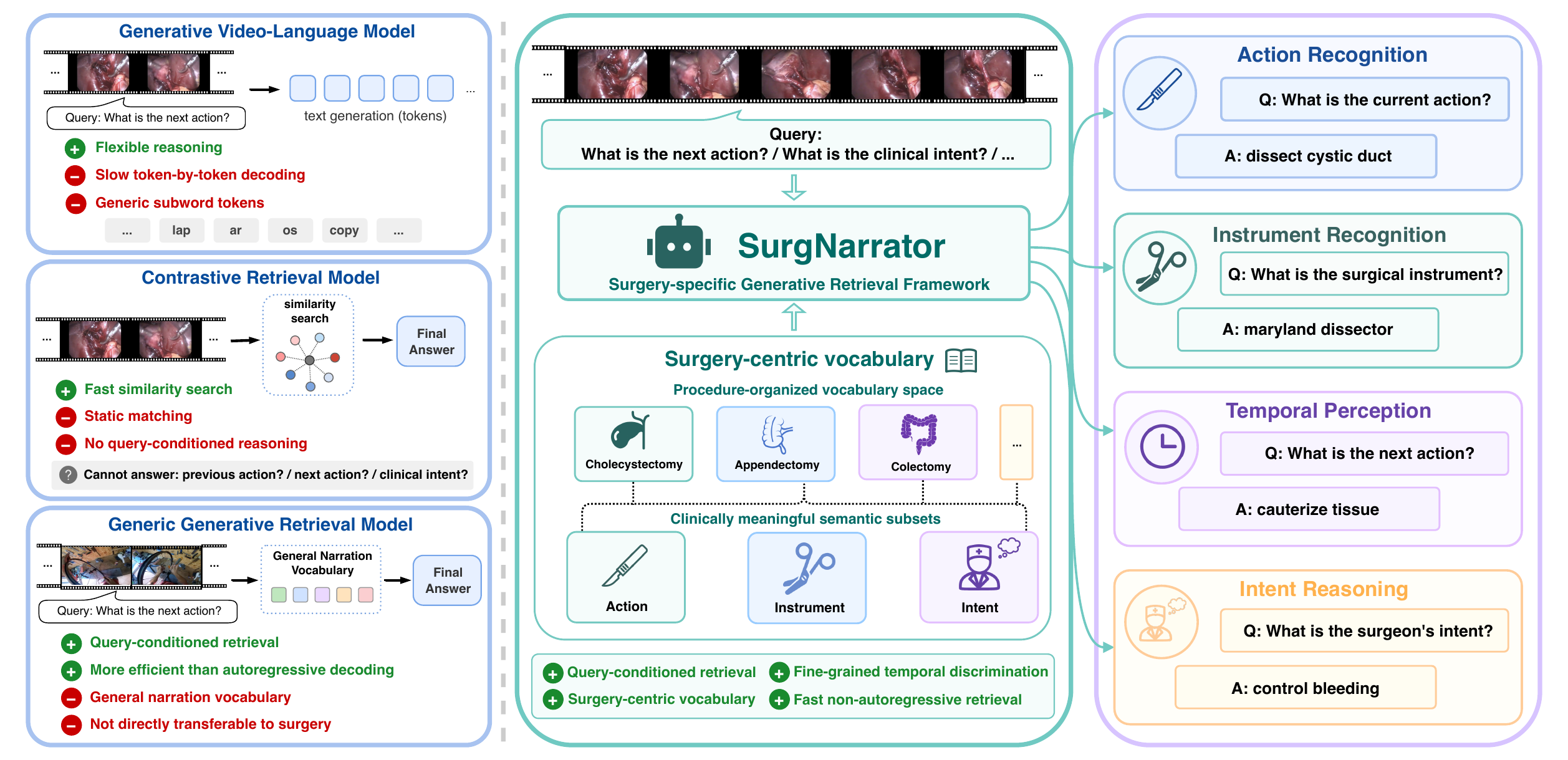}
\caption{Overview of SurgNarrator. SurgNarrator adapts generative retrieval to surgical video understanding by grounding retrieval in a surgery-centric vocabulary of clinical concepts. The vocabulary is organized by procedure type and semantic category, supporting tasks such as action recognition, instrument recognition, temporal perception, and intent reasoning without token-by-token autoregressive decoding.}
\label{figure1}
% \vspace{-10pt}
\end{figure*}

\IEEEPARstart{S}{urgical} video understanding aims to interpret intraoperative clinical events within evolving surgical workflows, supporting clinically meaningful analysis of complex procedures~\cite{Maier_Hein_2017,twinanda2016endonet}. This capability is essential for downstream applications such as AI copilots for live surgical assistance, postoperative analysis, and even trainee education~\cite{czempiel2020tecno,funke2019video,twinanda2016endonet,Maier_Hein_2017}. Typically, surgical videos follow structured and recurring sequences of clinical events, yet their interpretation depends on fine-grained temporal context and clinically meaningful distinctions~\cite{czempiel2020tecno,nwoye2023cholectriplet2021}. Therefore, models for this setting should deliver clinically faithful and interpretable predictions to provide intraoperative guidance in real time, rather than mere approximate descriptions.

Nevertheless, existing methods often impose a trade-off between effectiveness and efficiency. For generative video-language models, they can reason flexibly over open-ended queries~\cite{maaz2024video,zhang2023video,li2024llava}. However, this capability relies on autoregressive decoding, where answers are generated token-by-token, inevitably introducing inference latency. Moreover, these models typically rely on large general-purpose subword vocabularies. For example, LLaMA's vocabulary spans 128K subword units~\cite{grattafiori2024llama}, while Qwen3 adopts a byte-level byte-pair encoding tokenizer with more than 151K vocabulary entries~\cite{qwen3}. Such generic subword tokenization often lacks clinical interpretability, as fragmented tokens such as ``lap'', ``ar'', ``os'', and ``copy'' carry no meaningful surgical semantics on their own, which may increase the risk of model hallucination and compromise contextual accuracy. In contrast, for contrastive retrieval models that match video embeddings against candidate descriptions, inference is fast~\cite{LUO2022293,xu2021videoclip,yuan2025learning}. However, these models are typically based on simple similarity search and operate as static matchers with no query-conditioned reasoning. As a consequence, they are structurally limited in answering questions such as ``what action comes next?'' or ``what is the clinical intent of this step?''. 

Recently, generative retrieval has been proposed and shows promise in bridging these two paradigms. Concretely, it reformulates decoding as reasoning-aware retrieval over a compact, non-subword vocabulary. By retrieving from a narrower and more precise space, it can improve both efficiency and effectiveness in general video understanding~\cite{lin2025vlog}. Nevertheless, transferring such a method to the clinical domain remains challenging. Specifically, surgical workflows involve domain-specific clinical terminology that is often absent from general narration vocabularies, and they contain events that are visually nearly identical but semantically distinct. For example, consecutive clips from the same procedure often look almost the same while denoting different procedural steps. As a result, discriminative representations built on clinical concepts are preferred. 

In this paper, we present SurgNarrator, a new generative retrieval framework for surgical video understanding. As shown in Fig.~\ref{figure1}, SurgNarrator delivers query-conditioned reasoning at retrieval-level latency, enabling surgical reasoning tasks such as temporal and intent inference. To this end, it grounds retrieval in a surgery-centric vocabulary of clinical concepts extracted from operative captions and adapts a pre-trained multimodal embedding model, Qwen3-VL-Embedding-8B~\cite{li2026qwen3}, to the surgical domain via a temporally-aware contrastive objective. At inference, SurgNarrator applies a hierarchical, procedure-aware retrieval mechanism that integrates clinical prior knowledge to improve inference effectiveness. Our contributions are thus summarized as follows:
\begin{itemize}
\item We introduce SurgNarrator, a novel generative retrieval framework that reformulates surgical video understanding as query-conditioned retrieval over a structured clinical concept space.
\item We construct a surgery-centric clinical concept vocabulary that captures clinically meaningful semantics and generalizes effectively across diverse surgical understanding tasks.
\item We propose a temporally-aware contrastive fine-tuning strategy that learns temporally discriminative representations, improving the distinction between visually similar yet semantically different surgical events.
\item Extensive experiments across diverse surgical understanding tasks and 12 zero-shot downstream benchmarks demonstrate that SurgNarrator consistently outperforms state-of-the-art methods while reducing output-stage latency by over two orders of magnitude.
\end{itemize}

\section{Related Work}
\subsection{Surgical Vision-Language Pretraining and Retrieval}
Surgical vision-language pretraining has become an important direction for learning transferable representations from surgical videos paired with textual supervision. Yuan et al.~\cite{yuan2025learning} introduced SurgVLP, a multimodal representation learning framework that aligned surgical video clips with automatically generated speech transcripts from large-scale surgical video lectures. HecVL was introduced in~\cite{yuan2024hecvl} for hierarchical video-language pretraining using clip-level action transcripts, phase-level conceptual summaries, and video-level procedure abstracts. Honarmand et al.~\cite{vidlpro} presented VidLPRO, combining video-text contrastive learning, video-text matching, and masked language modeling for robotic and laparoscopic surgery. For ophthalmic surgery, Hu et al. developed OphCLIP, which leveraged large-scale silent surgical videos as a knowledge base for retrieval-augmented pretraining~\cite{hu2025ophclip}.

Several studies have also explicitly incorporated temporal information into surgical video-language learning, including temporal ordering and cross-clip context. Yuan et al. designed PeskaVLP with hierarchical knowledge augmentation, temporally reversed text sequences as hard negatives, and a Dynamic Time Warping objective for procedural alignment~\cite{yuan2024procedure}. Stilz et al.~\cite{stilz2026clipper} developed CliPPER for long-form intraoperative videos, using contextual video-text contrastive learning, cycle-consistency alignment, frame-text matching, and clip order prediction to model temporal dependencies in surgical events. These studies show that surgical video understanding benefits from language supervision, procedural hierarchy, temporal ordering, and cross-clip context. However, these approaches are mainly optimized for transferable representation learning and downstream recognition, with limited capacity for query-specific reasoning within complex surgical workflows.

\subsection{Generative Models for Surgical Video Understanding}
Generative vision-language models have increasingly been adapted to surgical vision-language interaction, particularly for surgical visual question answering (VQA) and conversational assistance. Seenivasan et al. proposed SurgicalGPT, which extended GPT-2 with a learnable visual tokenizer and visual token embeddings for end-to-end surgical visual question answering~\cite{seenivasan2023surgicalgpt}. Subsequently, Li et al. introduced LLaVA-Surg, a multimodal surgical assistant trained on 102K surgical video-instruction pairs generated from surgical lecture videos~\cite{li2024llava}. Schmidgall et al. presented GP-VLS, a general-purpose surgical vision-language model that integrated medical and surgical knowledge with visual scene understanding~\cite{schmidgall2024gp}. Hou et al. developed S$^2$CAN, a memory-augmented multimodal framework that used self-contained inquiry to construct direct and indirect memory for surgical VQA~\cite{hou2024memory}.

Recent studies have further advanced surgical generative models toward grounded perception, temporal modeling, and explicit reasoning. Wang et al.~\cite{WANG2026103789} introduced EndoChat, a grounded multimodal large language model for endoscopic surgery with a mixed visual token engine and a hallucination mitigation strategy. They also proposed Surgical-LVLM for grounded robotic surgery through Visual Perception LoRA blocks and a Token-Interaction module~\cite{wang2024surgical}. SurgVidLM was developed to combine global context with local temporal details for multi-grained surgical video understanding~\cite{wang2025surgvidlm}. Zeng et al.~\cite{zeng2025surgvlm} developed SurgVLM, a surgical vision-language foundation model for visual perception, temporal analysis, and high-level reasoning across diverse surgical tasks. Li et al. trained SurgLLaVA-Video, a surgical vision–language model that processed both frame- and video-level surgical inputs across multiple surgical specialties~\cite{Li_2026}. Perez et al.~\cite{perez2026sureon} proposed SureonVLM and SureonVLM-R1 for surgical reasoning, where the former was adapted through supervised fine-tuning and the latter was optimized with Group Relative Policy Optimization to answer reasoning-intensive questions grounded in surgical videos. Jiang et al. developed Surg-R1, a hierarchical surgical reasoning foundation model that decomposed surgical interpretation into perceptual grounding, relational understanding, and contextual reasoning through a four-stage training pipeline spanning supervised fine-tuning, reinforcement learning, and iterative self-improvement~\cite{jiang2026surg}.

Existing surgical generative models have advanced open-ended surgical reasoning, but they commonly rely on autoregressive decoding over large general-purpose subword vocabularies. Although this paradigm supports flexible natural-language interaction, sequential token generation increases inference latency and provides limited control over clinically structured output spaces.

\subsection{Generative Retrieval for Video Understanding}
Generative retrieval has recently been explored as an efficient alternative to fully autoregressive generation for video understanding. Lin et al.~\cite{lin2025vlog} proposed VLog, which organized video narrations into a vocabulary and retrieved narration units instead of generating outputs token by token. Built on GPT-2, VLog combined language-model reasoning with vocabulary-based retrieval, preserving query-conditioned prediction while reducing decoding latency. This work demonstrated that textual semantic units can support efficient video understanding without fully autoregressive decoding.

However, adapting such a framework to surgical video understanding introduces several unique challenges. Surgical workflows involve domain-specific clinical terminology that is absent from general-purpose narration vocabularies. Moreover, consecutive surgical clips frequently exhibit highly similar visual appearances despite corresponding to semantically distinct procedural events, making fine-grained temporal discrimination difficult for conventional contrastive learning objectives. Motivated by these challenges, this study introduces SurgNarrator, which performs generative retrieval with a surgery-centric vocabulary and temporally-aware contrastive fine-tuning for surgical video understanding.

\section{SurgNarrator}

\begin{figure*}[!t]
\centering
\includegraphics[width=1\textwidth]{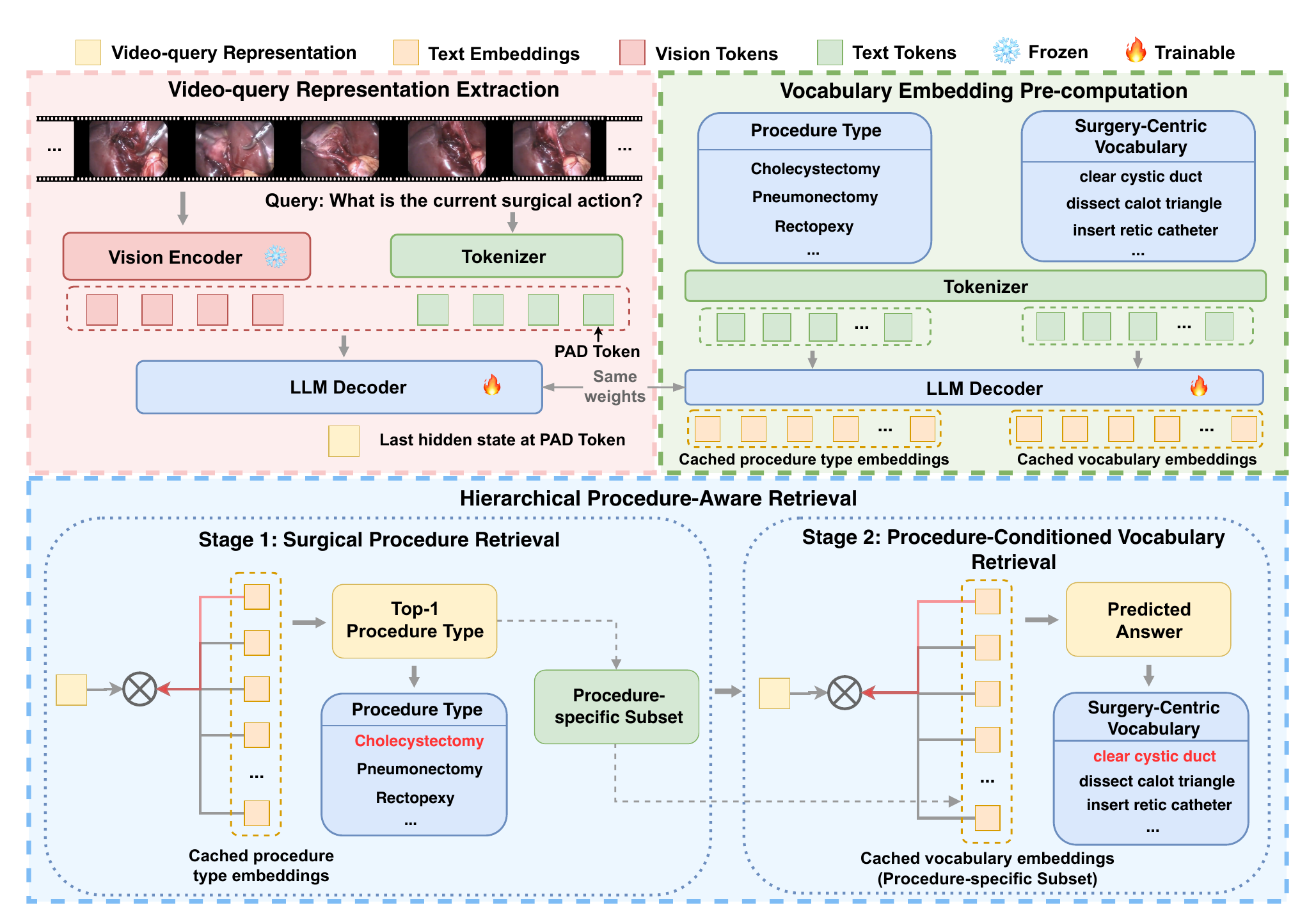}
\caption{Hierarchical procedure-aware generative retrieval pipeline of SurgNarrator. Vocabulary and procedure-type embeddings are pre-computed and cached for efficient inference. Given a surgical video-query pair, SurgNarrator obtains a dense video-query representation from the last hidden state of the appended PAD token. It first identifies the procedure type, then retrieves the answer from the corresponding procedure-specific vocabulary without autoregressive decoding.}
\label{figure2}
% \vspace{-10pt}
\end{figure*}

An overview of the SurgNarrator pipeline is illustrated in Fig.~\ref{figure2}. SurgNarrator reformulates surgical video understanding as query-conditioned retrieval over a structured clinical concept space, rather than directly generating answers with autoregressive large vision-language models. Given a surgical video-query pair, SurgNarrator retrieves the most relevant clinical concept through hierarchical procedure-aware retrieval. To this end, we introduce (1) a surgery-centric vocabulary to model clinically meaningful concepts, (2) temporally-aware contrastive learning to produce representations that preserve fine-grained surgical semantics and temporal progression, and (3) hierarchical procedure-aware retrieval to progressively reduce the search space for efficient and accurate answer prediction.

\subsection{Problem Definition and Notations}
For the $i$-th video-query instance, the surgical video clip is represented as an ordered sequence of sampled frames,
$V_i=[v_{i,1},v_{i,2},\ldots,v_{i,T}]$, where $v_{i,t}$ denotes the $t$-th sampled frame and $T$ is the total number of sampled frames. The associated natural-language query and textual answer are denoted by $Q_i$ and $a_i$, respectively. Each instance is associated with a procedure type $p_i\in\mathcal{P}$, a semantic annotation level $\ell_i\in\{\mathrm{coarse},\mathrm{mid},\mathrm{fine}\}$, a temporal position $\tau_i$, and a source video identity $u_i$. Here, $\ell_i$ specifies the hierarchy level used for temporal hard negative mining, while $\tau_i$ and $u_i$ are used to identify the temporal order of the clip within its source video.

Given $V_i$ and $Q_i$, SurgNarrator derives a video-query representation $\mathbf{e}_i$ and retrieves the most relevant answer from a surgery-centric vocabulary. The embedding for each candidate vocabulary entry $a_j\in\mathcal{A}$ is defined as:
\begin{equation}
\mathbf{o}_j=f_{\theta}(a_j),
\label{eq:vocabulary-embedding}
\end{equation}
where $f_{\theta}$ denotes the embedding function and
$\mathbf{o}_j$ denotes the representation of candidate
vocabulary entry $a_j$.

Let $\mathcal{A}_p\subseteq\mathcal{A}$ denote the vocabulary subset associated with procedure type $p\in\mathcal{P}$, and let $\Omega_i$ denote the retrieval space for the $i$-th instance. In global retrieval, $\Omega_i=\mathcal{A}$. In procedure-aware retrieval, $\Omega_i=\mathcal{A}_{\hat{p}_i}$, where $\hat{p}_i$ is the retrieved procedure type. The prediction is obtained by
\begin{equation}
\hat{a}_i =
\operatorname*{arg\,max}_{a_j\in\Omega_i}
s(\mathbf{e}_i,\mathbf{o}_j),
\label{eq:answer-retrieval}
\end{equation}
where $s(\cdot,\cdot)$ denotes the similarity function. This formulation constrains the output space to clinically meaningful surgical concepts while avoiding autoregressive decoding over a generic subword vocabulary.

\begin{figure*}[!t]
\centering
\includegraphics[width=1\textwidth]{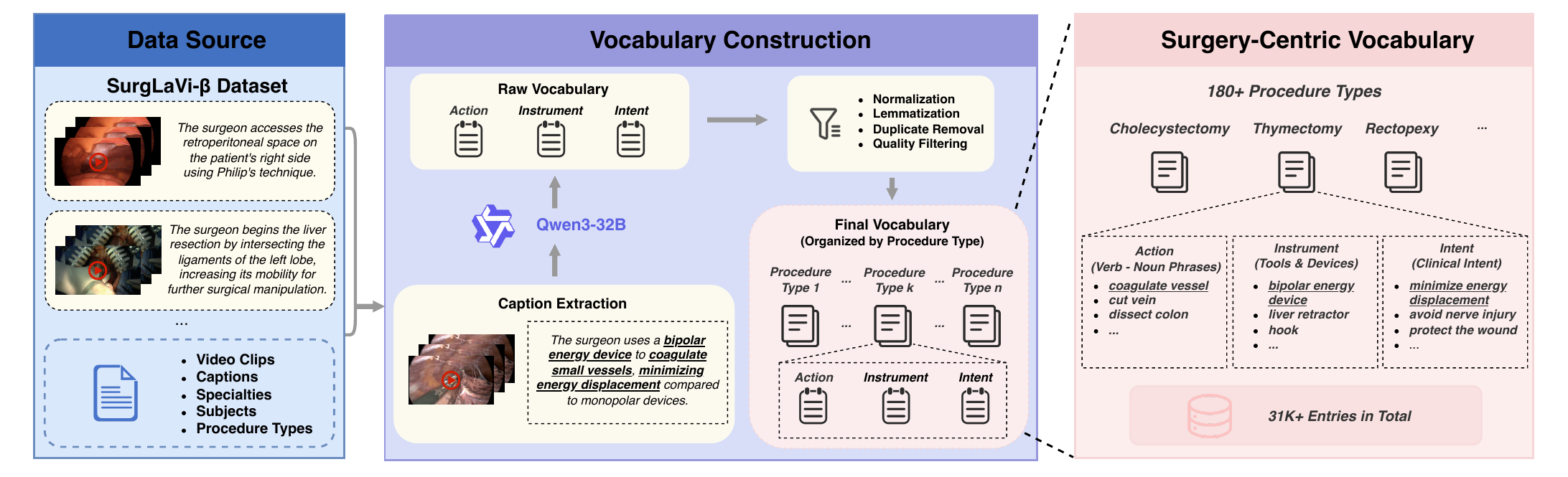}
\caption{The process of surgery-centric vocabulary construction. We employ Qwen3-32B to extract raw vocabulary entries from the captions along three dimensions: action, instrument, and intent. After careful filtering and deduplication, the resulting terms are organized by procedure type to constitute the final surgery-centric vocabulary.}
\label{figure3}
% \vspace{-10pt}
\end{figure*}

\subsection{Query-Conditioned Generative Retrieval}
Given a surgical video-query pair $(V_i,Q_i)$, SurgNarrator predicts the answer by retrieving the most relevant entry from the surgery-centric vocabulary. To generate video-query representations, we use an embedding model initialized from Qwen3-VL-Embedding-8B~\cite{li2026qwen3} and fine-tuned on surgical data.

Let $f_{\theta}$ denote the resulting embedding function. The dense video-query representation is obtained from the last hidden state of the appended PAD token. Specifically, the video-query pair is encoded as:
\begin{equation}
\mathbf{e}_i=f_{\theta}(V_i,Q_i),
\end{equation}
where $(V_i,Q_i)$ is a multimodal input consisting of sampled surgical frames and a task-specific query, and $\mathbf{e}_i$ denotes the video-query representation. 

As defined in~\eqref{eq:vocabulary-embedding}, the same embedding function is used to encode each text-only vocabulary entry $a_j\in\mathcal{A}$ into its representation $\mathbf{o}_j$. This function maps video-query inputs and vocabulary entries into the same representation space. The vocabulary embeddings $\mathbf{o}_j$ are pre-computed offline and cached, eliminating repeated vocabulary encoding during inference.

All embeddings are L2-normalized before similarity computation. The relevance between a video-query instance and a candidate vocabulary entry is measured as:
\begin{equation}
s(\mathbf{e}_i,\mathbf{o}_j)=\mathbf{e}_i^\top\mathbf{o}_j .
\end{equation}

SurgNarrator ranks candidate entries in $\Omega_i$ according to this similarity score and retrieves the top-ranked entry as defined in~\eqref{eq:answer-retrieval}. This retrieval formulation produces task-specific textual predictions from the surgery-centric vocabulary without autoregressive token generation.

\subsection{Surgery-Centric Vocabulary Construction}
The process of surgery-centric vocabulary construction is illustrated in Fig.~\ref{figure3}. The vocabulary is constructed from SurgLaVi-$\beta$~\cite{PEREZ2026103982}, which provides surgical video clips paired with descriptive captions and hierarchical annotations. These captions describe surgical actions, instruments, and procedural intents, making them a suitable source for constructing a structured surgery-centric vocabulary.

To extract structured surgical events from free-form captions, we use Qwen3-32B~\cite{qwen3}, a large language model with strong instruction-following capabilities. Each caption is decomposed into three semantic categories: surgical action, instrument, and clinical intent. Action entries are represented as verb--noun phrases, such as ``cut tissue.'' Instrument entries correspond to surgical tools and devices, whereas intent entries describe the clinical purpose underlying the observed activity, such as ``avoid nerve injury.'' To reduce hallucinated entries, the extraction is constrained to information explicitly stated or directly supported by the original caption. Accordingly, only caption-grounded actions, instruments, and clinical intents are retained as vocabulary candidates.

The raw entries are then standardized through a normalization pipeline. Specifically, punctuation marks are removed, casing is normalized, inflected word forms are lemmatized, and duplicate entries are merged. After filtering and deduplication, the final vocabulary contains more than 31K entries.

Beyond serving as a candidate set, the vocabulary defines the clinically meaningful output space of SurgNarrator. For each procedure type $p\in\mathcal{P}$, the procedure-specific vocabulary $\mathcal{A}_{p}$ is organized into three semantic subsets, including $\mathcal{A}_{p}^{\mathrm{act}}$ for surgical actions, $\mathcal{A}_{p}^{\mathrm{ins}}$ for instruments, and $\mathcal{A}_{p}^{\mathrm{int}}$ for clinical intents, where $\mathcal{A}_{p}=\mathcal{A}_{p}^{\mathrm{act}}\cup\mathcal{A}_{p}^{\mathrm{ins}}\cup\mathcal{A}_{p}^{\mathrm{int}}$. This caption-grounded structure preserves procedure-level organization and category-level surgical semantics, provides candidate concepts for both recognition- and reasoning-oriented tasks, and enables global retrieval over $\mathcal{A}$ as well as procedure-specific retrieval over $\mathcal{A}_{p}$.

\subsection{Temporally-Aware Contrastive Learning}
\begin{figure}[!t]
\centering
\includegraphics[width=\columnwidth]{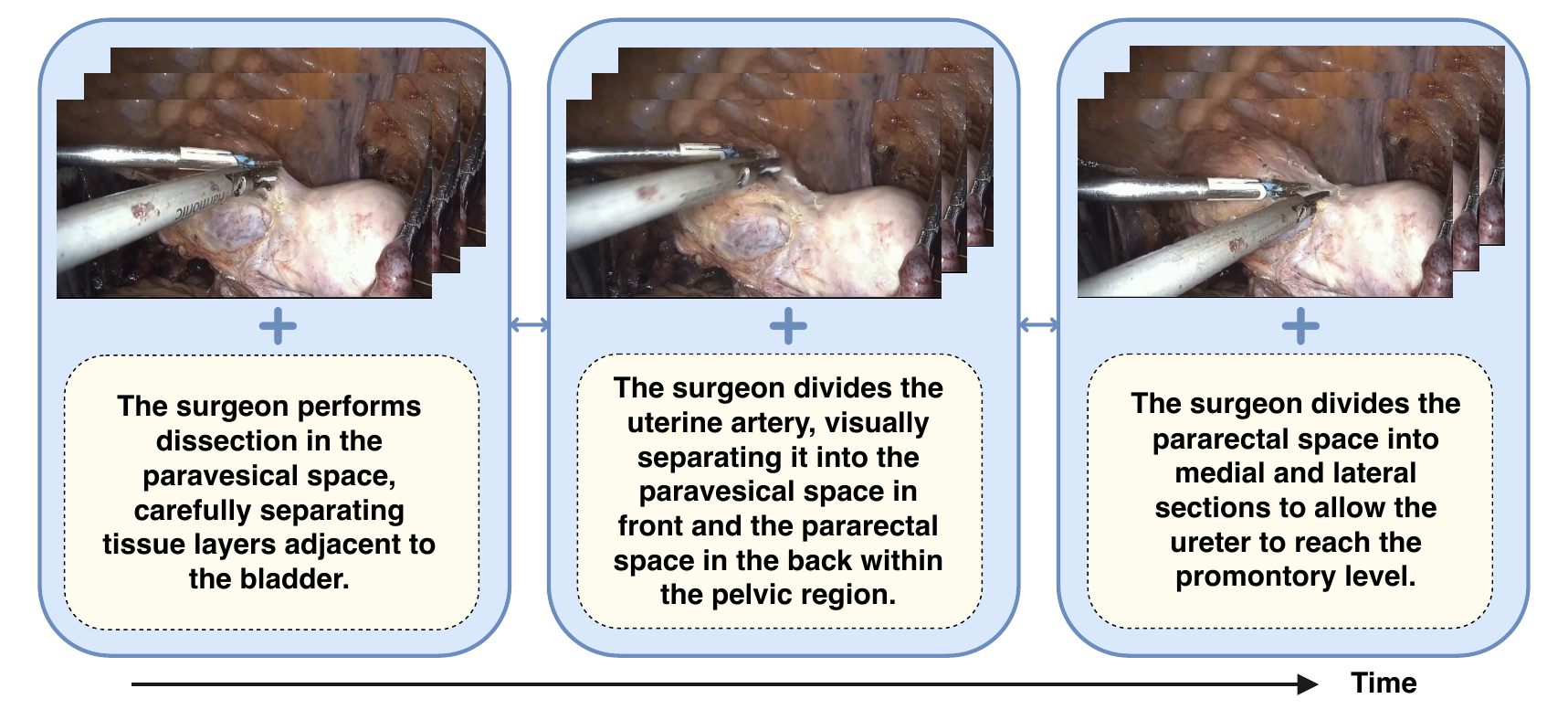}
\caption{Temporally adjacent surgical clips share a nearly identical visual appearance while corresponding to semantically distinct surgical events.}
\label{figure4}
% \vspace{-10pt}
\end{figure}

Standard contrastive learning aligns matched video-text pairs by treating non-matching samples within a batch as negatives. However, this assumption can be suboptimal for surgical videos. As shown in Fig.~\ref{figure4}, temporally adjacent clips often exhibit highly similar visual appearances while corresponding to semantically distinct procedural events, making them informative hard negatives for fine-grained temporal discrimination. At the same time, distinct clips may still correspond to overlapping surgical semantics, and directly treating them as negatives can introduce false-negative supervision. To address these challenges, we introduce a temporally-aware contrastive objective that incorporates temporally neighboring surgical clips as hard negative candidates while masking semantically overlapping samples.

\subsubsection{Temporal Hard Negative Mining}

Let $\mathcal{B}$ denote the batch index set used to estimate the contrastive objective, with $|\mathcal{B}|$ indicating the batch size. For each $i\in\mathcal{B}$, the positive pair consists of the video-query representation $\mathbf{e}_i=f_{\theta}(V_i,Q_i)$ and the corresponding answer representation $\mathbf{o}_i=f_{\theta}(a_i)$. The standard InfoNCE objective~\cite{oord2018representation} is formulated as
\begin{equation}
\mathcal{L}_{\mathrm{InfoNCE}}
=
-\frac{1}{|\mathcal{B}|}
\sum_{i\in\mathcal{B}}
\log
\frac{
\exp(s(\mathbf{e}_i,\mathbf{o}_i)/\gamma)
}{
\sum_{j\in\mathcal{B}}
\exp(s(\mathbf{e}_i,\mathbf{o}_j)/\gamma)
},
\end{equation}
where $\gamma$ is the temperature parameter. In the denominator, $\mathbf{o}_i$ is the positive answer representation for sample $i$, while $\{\mathbf{o}_j\}_{j\in\mathcal{B},j\ne i}$ are treated as negative candidates.

To exploit the temporal structure of surgical workflows, we define temporal hard negative candidates according to the source video identity, temporal position, and semantic annotation level. For sample $i$, the temporal hard negative set is defined as
\begin{equation}
\mathcal{N}_{\ell_i}(i)
=
\{j \mid u_j=u_i,\ \ell_j=\ell_i,\ 0<|\tau_j-\tau_i|\le r\}.
\end{equation}
Here, $r$ denotes the temporal neighborhood radius. In practice, $r=1$ selects the immediately preceding and following clips within the same source video and at the same semantic level. These neighboring clips are visually close to the anchor clip but may correspond to distinct procedural semantics, making them informative hard negatives for fine-grained surgical video retrieval.

\subsubsection{False Negative Masking and Optimization Objective}
Temporal hard negative mining enlarges the negative pool with temporally neighboring candidates that are useful for learning fine-grained procedural discrimination. However, some candidates may still exhibit semantic overlap with the positive answer because surgical annotations are hierarchical and repetitive. Treating such candidates as negatives may separate clinically related representations in the embedding space. We therefore mask potential false negatives by excluding candidates whose answer representations are highly similar to the positive answer.

For each anchor sample $i$, the candidate negative set $\mathcal{C}(i)$ is defined as:
\begin{equation}
\mathcal{C}(i)
=
(\mathcal{B}\setminus\{i\}) \cup \mathcal{N}_{\ell_i}(i),
\end{equation}
where $\mathcal{B}\setminus\{i\}$ denotes the negative set from the batch and $\mathcal{N}_{\ell_i}(i)$ denotes the temporal hard negative set.

A candidate sample is treated as a potential false negative when its answer representation is highly similar to the positive answer representation:
\begin{equation}
\mathcal{F}(i)
=
\{j\in\mathcal{C}(i)\mid s(\mathbf{o}_i,\mathbf{o}_j)\ge \delta\},
\end{equation}
where $\mathbf{o}_i$ is the positive answer representation, $\mathbf{o}_j$ is the answer representation of candidate $j$, and $\delta$ is the false negative masking threshold. 

The valid negative set is then given by
\begin{equation}
\mathcal{H}(i)
=
\mathcal{C}(i)\setminus\mathcal{F}(i).
\end{equation}

The final temporally-aware contrastive objective is formulated as
\begin{equation}
\mathcal{L}_{\mathrm{temp}}
=
-\frac{1}{|\mathcal{B}|}
\sum_{i\in\mathcal{B}}
\log
\frac{
\exp(s(\mathbf{e}_i,\mathbf{o}_i)/\gamma)
}{
\sum_{j\in\{i\}\cup\mathcal{H}(i)}
\exp(s(\mathbf{e}_i,\mathbf{o}_j)/\gamma)
}.
\end{equation}
Here, $\{i\}\cup\mathcal{H}(i)$ contains the positive index and the valid negative indices after temporal mining and false negative masking. This objective maintains the alignment between the anchor representation and its corresponding positive answer, while contrasting the anchor against temporally informative and semantically valid negative candidates. In our implementation, $\delta=0.80$ is used to balance false negative suppression and hard negative retention.

\subsection{Hierarchical Procedure-Aware Retrieval}
\label{sec:hierarchical_retrieval}
During inference, SurgNarrator exploits the procedural organization of the surgery-centric vocabulary. Given a video-query representation, the framework first retrieves the most relevant procedure type from the set of surgical procedures $\mathcal{P}$ and then searches the corresponding procedure-specific vocabulary for the final answer. This coarse-to-fine strategy constrains candidate answers to the relevant procedure context and supports efficient retrieval without autoregressive decoding.

For each procedure type $p\in\mathcal{P}$, we encode its textual procedure label $c_p$ as
\begin{equation}
\mathbf{r}_p=f_{\theta}(c_p),
\end{equation}
where $\mathbf{r}_p$ denotes the procedure embedding. All procedure embeddings are pre-computed and cached. Given the video-query representation $\mathbf{e}_i$, the predicted procedure type is obtained by
\begin{equation}
\hat{p}_i
=
\operatorname*{arg\,max}_{p\in\mathcal{P}}
s(\mathbf{e}_i,\mathbf{r}_p).
\end{equation}
The retrieval space is then restricted to the procedure-specific vocabulary:
\begin{equation}
\Omega_i=\mathcal{A}_{\hat{p}_i}.
\end{equation}
The final answer is retrieved by
\begin{equation}
\hat{a}_i
=
\operatorname*{arg\,max}_{a_j\in\Omega_i}
s(\mathbf{e}_i,\mathbf{o}_j).
\end{equation}

This hierarchical retrieval strategy reduces interference from procedurally irrelevant vocabulary entries, while cached procedure and vocabulary embeddings enable similarity-based inference without autoregressive decoding.

\section{Experiments}
In this section, we compare SurgNarrator with representative generative and retrieval-based baselines and assess its generalization across diverse surgical scenarios. We also describe the evaluation metrics and implementation details.

\subsection{Datasets}
\subsubsection{Training Dataset}
Surg-Train is constructed from SurgLaVi-$\beta$~\cite{PEREZ2026103982}. To avoid overlap between training and evaluation samples, the dataset is split at the video level. For each surgical procedure type, videos are assigned to the training and evaluation splits at an approximate ratio of 9:1. The split is further adjusted to ensure that the evaluation set covers all procedure types, subjects, and surgical specialties present in the dataset. This strategy results in a training split containing 86.6\% of the original SurgLaVi-$\beta$~\cite{PEREZ2026103982} videos, with the remaining videos used for evaluation. Variable-duration clips are then extracted from the videos in each split according to the caption start and end timestamps.

Based on the hierarchical annotations provided in SurgLaVi-$\beta$~\cite{PEREZ2026103982}, including coarse-, mid-,
and fine-level descriptions, we construct query-conditioned training samples for surgical video understanding. Overall, Surg-Train contains approximately 97.6K samples.

\begin{table*}[t]
\centering
\caption{Evaluation on recognition-oriented tasks. Decode time is reported in milliseconds per query. FT indicates whether the model is fine-tuned on Surg-Train. Values in parentheses indicate the decoding speedup relative to the generative baseline.}
\label{tab:recognition_retrieval}

\setlength{\tabcolsep}{2.5pt}

\resizebox{\textwidth}{!}{%
\begin{tabular}{l c | *{5}{c} | *{5}{c}}
\hline
\multirow{2}{*}{Model}
& \multirow{2}{*}{FT}
& \multicolumn{5}{c|}{Instrument Recognition}
& \multicolumn{5}{c}{Action Recognition} \\
%\cline{3-12}

&
& Recall@1 & Recall@5 & Recall@10 & CIDEr & Decode Time
& Recall@1 & Recall@5 & Recall@10 & CIDEr & Decode Time \\
\hline

Qwen3-VL-8B-Instruct~\cite{Qwen3-VL}
& $\checkmark$
& 0.29 & 1.87 & 3.84 & 0.22 & 349.61
& 1.31 & 2.08 & 3.01 & 0.15 & 394.44 \\

SurgCLIP-$\beta$~\cite{PEREZ2026103982}
& $\times$
& 5.57 & 16.95 & 22.30 & 33.94 & 56.83
& 1.65 & 6.87 & 10.57 & 30.95 & 58.50 \\

Qwen3-VL-Embedding-8B~\cite{li2026qwen3}
& $\times$
& 1.04 & 3.88 & 6.57 & 6.16 & 0.02
& 0.52 & 1.44 & 2.32 & 11.30 & 0.60 \\

SurgNarrator (Ours)
& $\checkmark$
& \textbf{7.11} & \textbf{17.81} & \textbf{23.81} & \textbf{38.57} & \textbf{0.02} ($\sim\mathbf{10^{4}\times}$)
& \textbf{2.30} & \textbf{8.34} & \textbf{11.77} & \textbf{33.43} & \textbf{1.17 ($\sim\mathbf{10^{2}\times}$)} \\

\hline
\end{tabular}%
}
\end{table*}

\begin{table*}[t]
\centering
\caption{Evaluation on reasoning-oriented tasks. Decode time is reported in milliseconds per query. FT indicates whether the model is fine-tuned on Surg-Train. ``--'' indicates the model cannot perform the corresponding task. Values in parentheses indicate the decoding speedup relative to the generative baseline.}
\label{tab:reasoning_retrieval}

\setlength{\tabcolsep}{2.5pt}

\resizebox{\textwidth}{!}{%
\begin{tabular}{l c | *{5}{c} | *{5}{c}}
\hline
\multirow{2}{*}{Model}
& \multirow{2}{*}{FT}
& \multicolumn{5}{c|}{Temporal Perception}
& \multicolumn{5}{c}{Intent Reasoning} \\
%\cline{3-12}

&
& Recall@1 & Recall@5 & Recall@10 & CIDEr & Decode Time
& Recall@1 & Recall@5 & Recall@10 & CIDEr & Decode Time \\
\hline

Qwen3-VL-8B-Instruct~\cite{Qwen3-VL}
& $\checkmark$
& 0.67 & 1.29 & 1.91 & 0.10 & 197.35
& 0.34 & 1.89 & 3.96 & 0.05 & 409.85 \\

SurgCLIP-$\beta$~\cite{PEREZ2026103982}
& $\times$
& -- & -- & -- & -- & --
& -- & -- & -- & -- & -- \\

Qwen3-VL-Embedding-8B~\cite{li2026qwen3}
& $\times$
& 0.32 & 1.12 & 1.89 & 8.37 & 1.07
& 0.77 & 9.64 & 11.62 & 10.41 & 0.34 \\

SurgNarrator (Ours)
& $\checkmark$
& \textbf{1.40} & \textbf{5.49} & \textbf{8.08} & \textbf{24.71} & \textbf{1.26 ($\sim\mathbf{10^{2}\times}$)}
& \textbf{2.07} & \textbf{9.94} & \textbf{15.96} & \textbf{19.28} & \textbf{0.77 ($\sim\mathbf{10^{2}\times}$)} \\

\hline
\end{tabular}%
}
\end{table*}

\subsubsection{Evaluation Datasets}
Surg-Eval is constructed from the evaluation split of SurgLaVi-$\beta$~\cite{PEREZ2026103982}, with no overlap with the training videos. It comprises four query-conditioned tasks grouped into recognition- and reasoning-oriented categories. The recognition-oriented tasks include instrument recognition and action recognition, which identify the instrument and surgical action present in the current video clip, respectively. The reasoning-oriented tasks include temporal perception, which identifies the previous, current, or next surgical action as specified by the query, and intent reasoning, which infers the surgeon's clinical intent.

In addition to Surg-Eval, SurgNarrator is evaluated on multiple downstream surgical benchmarks in a zero-shot setting. These benchmarks span diverse surgical procedures, modalities, and task settings, including phase, step, action, triplet, and tool recognition. Phase recognition is evaluated on Cholec80~\cite{twinanda2016endonet}, AutoLaparo~\cite{wang2022autolaparo}, StrasBypass70, BernBypass70~\cite{lavanchy2024challenges}, HeiChole~\cite{wagner2023comparative}, and GraSP~\cite{AYOBI2025103726}. The remaining tasks include step recognition on GraSP~\cite{AYOBI2025103726}, action recognition on SAR-RARP50~\cite{psychogyios2023sar}, triplet recognition on CholecT50~\cite{nwoye2022rendezvous}, and tool recognition on Cholec80~\cite{twinanda2016endonet}, HeiChole~\cite{wagner2023comparative}, and GraSP~\cite{AYOBI2025103726}. These evaluations assess the generalization capability of SurgNarrator across heterogeneous surgical video understanding scenarios beyond Surg-Eval.

\subsection{Evaluation Metrics}
For Surg-Eval, we evaluate both retrieval-based and autoregressive generative methods using the same set of metrics. We report Recall@K for vocabulary-level retrieval accuracy, with \(K \in \{1,5,10\}\), and CIDEr~\cite{vedantam2015cider} for textual quality. Recall@K indicates whether the ground-truth answer appears among the top-\(K\) candidates from the surgery-centric vocabulary. Retrieval-based methods obtain the candidate ranking directly by computing similarities between the video-query representation and vocabulary entries. For autoregressive generative baselines, we map each generated answer to a vocabulary ranking by computing cosine similarities to all candidate entries using all-MiniLM-L6-v2, a lightweight sentence transformer model~\cite{reimers2019sentence}. Recall@K is then computed on this mapped ranking. For textual quality evaluation, CIDEr is computed using the top-ranked retrieved answer for retrieval-based methods and the generated answer for autoregressive baselines. We also report the average decoding time per query to measure generation speed.

For zero-shot generalization on downstream surgical video understanding benchmarks, we follow the standard evaluation protocol for each task. For phase, step, and action recognition, we report video-wise accuracy and F1 score. For triplet and tool recognition, we report mean average precision (mAP).

\subsection{Implementation Details}
SurgNarrator is built on Qwen3-VL-Embedding-8B~\cite{li2026qwen3}, which serves as the multimodal embedding backbone. We uniformly sample 16 frames from each video clip and process them with the backbone's native visual encoder. We fine-tune the model using Low-Rank Adaptation (LoRA)~\cite{hu2022lora} for parameter-efficient adaptation to the surgical domain.

Training is conducted on eight NVIDIA A100 GPUs for 4 epochs with a batch size of 80. We use a learning rate of \(1\times10^{-4}\), a warmup ratio of 0.05, and a weight decay of 0.02. The contrastive temperature is set to \(\gamma=0.07\), and the false-negative masking threshold is set to \(\delta=0.80\).

For Surg-Eval, procedure and vocabulary embeddings are pre-computed and cached, following the hierarchical retrieval strategy described in Section~\ref{sec:hierarchical_retrieval}.

For downstream zero-shot evaluation, we follow the evaluation protocol of SurgLaVi-$\beta$~\cite{PEREZ2026103982}. Each evaluation sample is represented by a 16-frame temporal window centered on the evaluation frame. The resulting video-level embedding is compared with candidate class embeddings using cosine similarity. No task-specific fine-tuning is performed during zero-shot evaluation.

\begin{table*}[t]
\centering
\caption{Zero-shot phase and step recognition results. The best results are shown in bold, and the second-best results are underlined.}
\label{tab:phase_step}

\setlength{\tabcolsep}{3pt}
\renewcommand{\arraystretch}{1.05}

\begin{tabular}{l|cccccccccccc|cc|cc}
\hline

\multirow{3}{*}{Model}
& \multicolumn{12}{c|}{Phase Recognition}
& \multicolumn{2}{c|}{Step Recognition}
& \multicolumn{2}{c}{Avg} \\

& \multicolumn{2}{c}{Cholec80}
& \multicolumn{2}{c}{AutoLaparo}
& \multicolumn{2}{c}{StrasBypass70}
& \multicolumn{2}{c}{BernBypass70}
& \multicolumn{2}{c}{GraSP}
& \multicolumn{2}{c|}{HeiChole}
& \multicolumn{2}{c|}{GraSP}
& \multicolumn{2}{c}{} \\

\cline{2-17}

& Acc & F1
& Acc & F1
& Acc & F1
& Acc & F1
& Acc & F1
& Acc & F1
& Acc & F1
& Acc & F1 \\

\hline

CLIP~\cite{radford2021learning}
& 27.81 & 8.42
& 8.02 & 4.79
& 18.52 & 3.51
& 20.73 & 4.17
& 9.93 & 2.06
& 19.21 & 6.17
& 3.85 & 0.83
& 15.44 & 4.28 \\

MedSigLIP~\cite{sellergren2025medgemma}
& 40.94 & 18.74
& 26.54 & 10.17
& 20.15 & 5.54
& 30.73 & 6.14
& 12.88 & 4.84
& 45.36 & 17.63
& 11.53 & 1.86
& 26.88 & 9.27 \\

SurgVLP~\cite{yuan2025learning}
& 49.02 & 32.46
& 10.02 & 7.19
& 26.08 & 19.18
& 30.46 & \underline{17.48}
& 11.79 & 6.07
& 45.70 & 27.04
& 6.02 & 3.22
& 25.58 & 16.09 \\

HecVL~\cite{yuan2024hecvl}
& 47.50 & 23.89
& 38.72 & 25.81
& 28.59 & 23.91
& \textbf{32.09} & \textbf{18.58}
& 10.99 & 6.46
& 29.91 & 22.76
& 1.96 & 0.56
& 27.11 & 17.42 \\

PeskaVLP~\cite{yuan2024procedure}
& 51.43 & 39.39
& 34.94 & 28.41
& 29.42 & 19.64
& \underline{31.48} & 17.47
& 13.85 & 9.86
& 53.33 & 36.46
& 3.65 & 2.47
& 31.16 & 21.96 \\

\hline

SurgCLIP-$\beta$~\cite{PEREZ2026103982}
& \underline{57.98} & \underline{39.42}
& \underline{55.72} & \underline{45.95}
& \textbf{31.24} & \underline{26.05}
& 18.30 & 15.06
& \underline{34.77} & \underline{27.98}
& \underline{56.95} & \underline{44.00}
& \underline{14.15} & \underline{11.14}
& \underline{38.44} & \underline{29.94} \\

\rowcolor[gray]{0.875}
SurgNarrator (Ours)
& \textbf{63.60} & \textbf{48.72}
& \textbf{64.51} & \textbf{52.53}
& \underline{30.52} & \textbf{28.58}
& 20.66 & 17.39
& \textbf{39.76} & \textbf{32.12}
& \textbf{63.34} & \textbf{56.33}
& \textbf{29.05} & \textbf{18.77}
& \textbf{44.49} & \textbf{36.35} \\

\hline
\end{tabular}
\end{table*}

\subsection{Comparison With Generative and Retrieval Baselines}

Table~\ref{tab:recognition_retrieval} shows the recognition-oriented results on Surg-Eval, covering instrument and action recognition, whereas Table~\ref{tab:reasoning_retrieval} reports the reasoning-oriented results, covering temporal perception and intent reasoning. We compare SurgNarrator with Qwen3-VL-8B-Instruct~\cite{Qwen3-VL}, SurgCLIP-$\beta$~\cite{PEREZ2026103982}, and Qwen3-VL-Embedding-8B~\cite{li2026qwen3}. For a fair comparison, Qwen3-VL-8B-Instruct~\cite{Qwen3-VL} is fine-tuned using the same Surg-Train split used to adapt Qwen3-VL-Embedding-8B~\cite{li2026qwen3} in SurgNarrator, which corresponds to 86.6\% of the original SurgCLIP-$\beta$~\cite{PEREZ2026103982} pretraining data.

On recognition-oriented tasks, SurgNarrator outperforms both generative and retrieval baselines while maintaining retrieval-level inference efficiency. Compared with the autoregressive baseline, SurgNarrator achieves substantially better retrieval accuracy and semantic answer quality, while reducing output-stage latency by more than two orders of magnitude.

On reasoning-oriented tasks, SurgNarrator's advantage becomes more evident, despite lower absolute scores than those on recognition-oriented tasks. This is because these tasks require fine-grained procedural context rather than direct visual-category alignment. SurgCLIP-$\beta$~\cite{PEREZ2026103982} cannot directly address these tasks in its original retrieval formulation, as they require query-conditioned understanding of temporal context and procedural intent. In contrast, SurgNarrator remains applicable to both tasks and substantially improves over the two applicable baselines. These results demonstrate that transferring generative retrieval to surgical video understanding enables efficient, surgery-specific, and query-conditioned understanding across both tasks.

\begin{table}[t]
\centering
\caption{Zero-shot action and triplet recognition results. Bold and underlined values indicate the best and second-best results, respectively.}
\label{tab:action_triplet}

\renewcommand{\arraystretch}{1.05}
\setlength{\tabcolsep}{6pt}
%\small

\begin{tabular}{l|cc|c}
\hline

\multirow{3}{*}{Model}
& \multicolumn{2}{c|}{Action Recognition}
& \multicolumn{1}{c}{Triplet Recognition} \\

& \multicolumn{2}{c|}{SAR-RARP50}
& CholecT50 \\

\cline{2-4}

& Acc & F1 & mAP \\
\hline

CLIP~\cite{radford2021learning}
& \textbf{28.80}
& 6.28
& 2.50 \\

MedSigLIP~\cite{sellergren2025medgemma}
& 5.49
& 3.17
& 2.98 \\

SurgVLP~\cite{yuan2025learning}
& 17.91
& 7.03
& 3.04 \\

HecVL~\cite{yuan2024hecvl}
& 4.67
& 3.03
& 3.63 \\

PeskaVLP~\cite{yuan2024procedure}
& 5.18
& 2.68
& \underline{4.49} \\

\hline

SurgCLIP-$\beta$~\cite{PEREZ2026103982}
& 13.94
& \underline{7.62}
& 4.17 \\

\rowcolor[gray]{0.875}
SurgNarrator (Ours)
& \underline{20.64}
& \textbf{9.82}
& \textbf{5.65} \\

\hline
\end{tabular}
\end{table}

\subsection{Comparison on Generalization Capability}

To evaluate the generalizability of SurgNarrator, we further assess its transferability across 12 zero-shot surgical downstream tasks, including phase, step, action, triplet, and tool recognition. Table~\ref{tab:phase_step} reports the results for phase and step recognition, Table~\ref{tab:action_triplet} reports the results for action and triplet recognition, and Table~\ref{tab:tool} reports the results for tool recognition. Notably, SurgNarrator uses the same SurgLaVi-$\beta$~\cite{PEREZ2026103982} data source as SurgCLIP-$\beta$~\cite{PEREZ2026103982}, but is trained only on the Surg-Train split, corresponding to 86.6\% of the original SurgCLIP-$\beta$~\cite{PEREZ2026103982} pretraining data.

Despite using less training data than SurgCLIP-$\beta$~\cite{PEREZ2026103982}, SurgNarrator outperforms prior surgical vision-language models on most downstream benchmarks. The lower performance on BernBypass70~\cite{lavanchy2024challenges} is mainly attributed to differences in the training data used by the compared methods. These results suggest that SurgNarrator learns transferable surgical video representations that generalize across diverse zero-shot tasks.

\begin{table}[t]
\centering
\caption{Zero-shot tool recognition results. Bold and underlined values indicate the best and second-best results, respectively.}
\label{tab:tool}

\renewcommand{\arraystretch}{1.05}
\setlength{\tabcolsep}{6pt}
%\small

\begin{tabular}{l|ccc|c}
\hline

\multirow{3}{*}{Model}
& \multicolumn{3}{c|}{Tool Recognition} \\

& Cholec80
& GraSP
& HeiChole
& Avg \\

\cline{2-5}

& mAP
& mAP
& mAP
& mAP \\

\hline

CLIP~\cite{radford2021learning}
& 18.44
& 36.00
& 21.18
& 25.21 \\

MedSigLIP~\cite{sellergren2025medgemma}
& 17.92
& 37.18
& 20.58
& 25.23 \\

SurgVLP~\cite{yuan2025learning}
& 31.19
& 36.93
& 22.46
& 30.19 \\

HecVL~\cite{yuan2024hecvl}
& 25.11
& 36.75
& 19.36
& 27.07 \\

PeskaVLP~\cite{yuan2024procedure}
& \underline{38.88}
& 41.05
& \underline{32.82}
& \underline{37.58} \\

\hline

SurgCLIP-$\beta$~\cite{PEREZ2026103982}
& 36.77
& \underline{43.06}
& 31.47
& 37.10 \\

\rowcolor[gray]{0.875}
SurgNarrator (Ours)
& \textbf{44.92}
& \textbf{60.86}
& \textbf{45.75}
& \textbf{50.51} \\

\hline
\end{tabular}
\end{table}

\begin{figure*}[!t]
\centering
\includegraphics[width=0.97\textwidth]{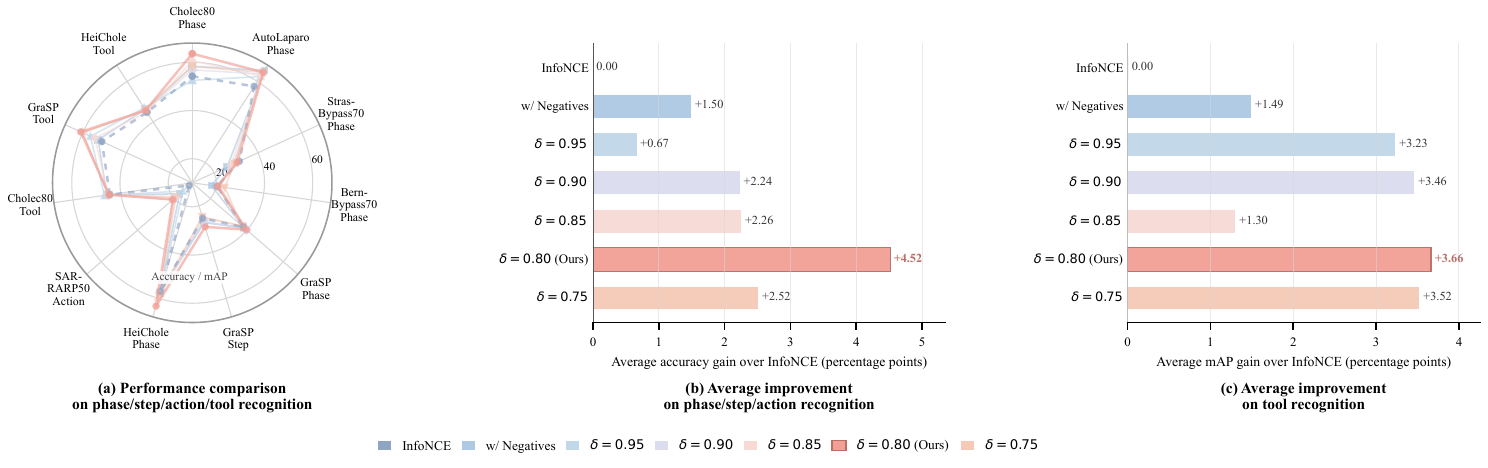}
\caption{Ablation study of temporally-aware hard negative sampling. We compare standard InfoNCE, temporally adjacent hard negatives without false-negative masking (\emph{w/ Negatives}), and different similarity thresholds for false-negative masking.
The threshold $\delta=0.80$ provides the best overall trade-off between suppressing semantically overlapping false negatives and preserving informative hard negatives.
}
\label{figure6}
% \vspace{-10pt}
\end{figure*}

\subsection{Ablation Study}
\subsubsection{Ablation of Temporally-Aware Hard Negative Sampling}
We further investigate the effectiveness of the proposed temporally-aware hard negative sampling strategy and the similarity-based false-negative masking mechanism on phase, step, action, and tool recognition benchmarks. As shown in Fig.~\ref{figure6}, we compare standard InfoNCE with a variant that adds temporally adjacent hard negatives without masking (\emph{w/ Negatives}) and variants using different masking thresholds $\delta$.

Introducing temporally adjacent hard negatives improves performance over standard InfoNCE, indicating that visually similar neighboring clips provide useful supervision for fine-grained procedural discrimination. The gain remains limited without false-negative masking, as the expanded negative pool may contain semantically overlapping candidates that are not reliable negatives.

Among different masking thresholds, $\delta=0.80$ achieves the best overall trade-off. It achieves the highest average accuracy gain on phase, step, and action recognition, improving over InfoNCE by 4.52 percentage points, and also achieves the highest average mAP gain on tool recognition. Larger thresholds such as $\delta=0.95$ and $\delta=0.90$ apply weaker masking and may still leave potential false negatives in the negative pool, whereas a smaller threshold such as $\delta=0.75$ may remove informative hard negatives. These results validate the importance of balancing hard negative mining with false-negative suppression.

\subsubsection{Effectiveness of Hierarchical Procedure-Aware Retrieval}

\begin{figure}[!t]
\centering
\includegraphics[width=\columnwidth]{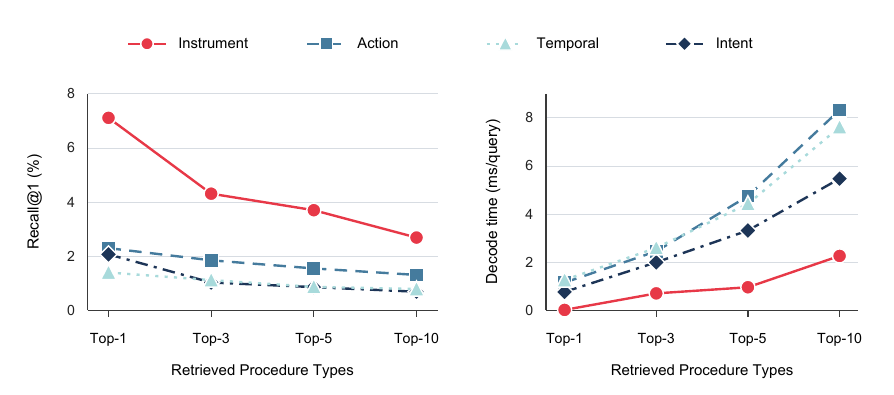}
\caption{Effect of the procedure-type retrieval space on retrieval accuracy and inference efficiency. Enlarging the procedure-type retrieval space consistently reduces Recall@1 while increasing decoding time across all tasks.}
\label{figure5}
% \vspace{-10pt}
\end{figure}

To evaluate the effectiveness of the proposed hierarchical
procedure-aware retrieval strategy, we compare different inference settings on Surg-Eval. In the hierarchical setting,
SurgNarrator first localizes the most relevant surgical procedure type and then performs answer retrieval within the corresponding procedure-specific vocabulary subset. We further
compare this design with direct retrieval from the full surgery-centric vocabulary and with variants that expand the second-stage retrieval space to the top-\(K\) retrieved procedure types.

Table~\ref{tab:ablation_retrieval} summarizes the overall ablation results. The top-1 procedure setting performs best across all tasks, showing that procedure-aware localization is important for accurate answer retrieval. In contrast, direct retrieval from the full vocabulary removes this localization constraint and forces each query to compete with candidate answers from heterogeneous surgical procedures. This increases cross-procedure semantic ambiguity and weakens retrieval precision. Similarly, including multiple procedure types introduces additional context-irrelevant vocabularies, leading to lower performance than the top-1 setting.

Fig.~\ref{figure5} further illustrates the accuracy-efficiency trade-off under different top-\(K\) procedure retrieval settings. As \(K\) increases, the retrieval space expands from a single procedure-specific vocabulary to the union of multiple procedure-specific vocabularies. Enlarging the retrieval space consistently reduces Recall@1 across all tasks, while increasing per-query decoding time because more candidate answers must be compared during retrieval. These results show that top-1 procedure localization provides the most favorable trade-off between retrieval precision and inference efficiency.

\begin{table}[t]
\centering
\caption{Ablation study of hierarchical procedure-aware retrieval on Surg-Eval. Recall@1 is reported for all tasks.}
\label{tab:ablation_retrieval}

\renewcommand{\arraystretch}{1.05}
\setlength{\tabcolsep}{5pt}
%\small

\begin{tabular}{l|cccc}
\hline

Retrieval Strategy
& Instrument
& Action
& Temporal
& Intent \\

\hline

Global
& 0.68
& 0.99
& 0.53
& 0.52 \\

Top-5 Procedures
& 3.70
& 1.55
& 0.87
& 0.86 \\

Top-3 Procedures
& 4.31
& 1.85
& 1.12
& 1.03 \\

Top-1 Procedure
& \textbf{7.11}
& \textbf{2.30}
& \textbf{1.40}
& \textbf{2.07} \\

\hline
\end{tabular}
\end{table}

\section{Conclusion}
In this work, we present SurgNarrator, a generative retrieval framework for surgical video understanding. We construct a surgery-centric vocabulary organized by procedure type and adapt Qwen3-VL-Embedding-8B with a temporally-aware contrastive fine-tuning strategy to learn discriminative surgical representations. We further introduce hierarchical procedure-aware retrieval to localize the candidate answer space and support efficient query-conditioned inference. Extensive experiments on Surg-Eval and twelve zero-shot downstream tasks demonstrate that SurgNarrator improves both recognition- and reasoning-oriented performance while maintaining retrieval-level inference efficiency. These results highlight the effectiveness and transferability of SurgNarrator for efficient surgical video understanding.

\section*{REFERENCES}
\bibliographystyle{IEEEtran}
\bibliography{references}

\end{document}